\documentclass{article}

\usepackage{arxiv}

\usepackage[american]{babel}
\usepackage[title]{appendix}
\usepackage{caption,subcaption}
\usepackage{graphicx}
\usepackage{algorithm}
\usepackage{algorithmic}
\usepackage{amsthm}
\usepackage{subdef}
\usepackage{wrapfig}
\usepackage[dvipsnames]{xcolor}

\usepackage{todonotes}
\usepackage{natbib} 
\usepackage{mathtools} 
\usepackage{booktabs} 
\usepackage{tikz} 

\usepackage{multirow}
\usepackage{adjustbox}
\usepackage{caption}
\usepackage{colortbl}

\usepackage[utf8]{inputenc} 
\usepackage[T1]{fontenc}    
\usepackage{hyperref}       
\usepackage{url}            
\usepackage{booktabs}       
\usepackage{amsfonts}       
\usepackage{nicefrac}       
\usepackage{microtype}      
\usepackage{lipsum}
\usepackage{times}
\usepackage{epsfig}
\usepackage{graphicx}
\usepackage{amsmath}
\usepackage{amssymb}
\usepackage{multirow}
\usepackage{soul}
\usepackage{float}

\newcommand{\modelss}{\mbox{\textsc{TSS}}}

\newcommand*\samethanks[1][\value{footnote}]{\footnotemark[#1]}

\title{Submodular Mutual Information for Targeted Data Subset Selection}

\author{Suraj Kothawade \thanks{Department of Computer Science, University of Texas at Dallas}\\
\texttt{suraj.kothawade@utdallas.edu} \\
\And
Vishal Kaushal \thanks{Department of Computer Science and Engineering, Indian Institute of Technology Bombay}\\
\texttt{vkaushal@cse.iitb.ac.in} \\
\And
Ganesh Ramakrishnan \samethanks \\
\texttt{ganesh@cse.iitb.ac.in} \\
\And
Jeff Bilmes \thanks{Department of Electrical \& Computer Engineering, University of Washington, Seattle}\\
\texttt{bilmes@uw.edu} \\
\And
Rishabh Iyer \samethanks[1]\\
\texttt{rishabh.iyer@utdallas.edu} \\
}

\begin{document}

\begin{center}
    \maketitle
\end{center}

\begin{abstract}
With the rapid growth of data, it is becoming increasingly difficult to train or improve deep learning models with the right subset of data. We show that this problem can be effectively solved at an additional labeling cost by \textit{targeted data subset selection} (\modelss) where a subset of unlabeled data points similar to an auxiliary set are added to the training data. We do so by using a rich class of Submodular Mutual Information (SMI) functions and demonstrate its effectiveness for image classification on CIFAR-10 and MNIST datasets. Lastly, we compare the performance of SMI functions for \modelss \vspace{0.5pt} with other state-of-the-art methods for closely related problems like active learning. Using SMI functions, we observe \textbf{$\approx$ 20-30\% gain over the model's performance before re-training with added targeted subset; $\approx$ 12\% more than other methods}.
\end{abstract}

\section{Introduction}
Recent times have seen unprecedented growth in data across modalities such as text, images and videos. This has naturally given rise to techniques for finding effective smaller subsets of the data for a variety of end-tasks. An example of this is data subset selection for efficient and/or cost-effective training of machine learning models, wherein we need to select samples which are most informative for training a model. Training on such smaller subsets of data often entails significant speedups and reduction in labeling time/cost without sacrificing much on accuracy~\citep{killamsetty2020glister,kaushal2019learning,wei2015submodularity}. Another flavor of this is targeted data subset selection which focuses on improving an existing model which is performing poorly is specific cases or improving a dataset which is imbalanced in certain attributes. Quite often, in these end-tasks, we want to be able to select {\em subsets that align well with a certain target set}. 

\begin{wrapfigure}{R}{0.50\textwidth}
\centering
\includegraphics[width=0.50\textwidth]{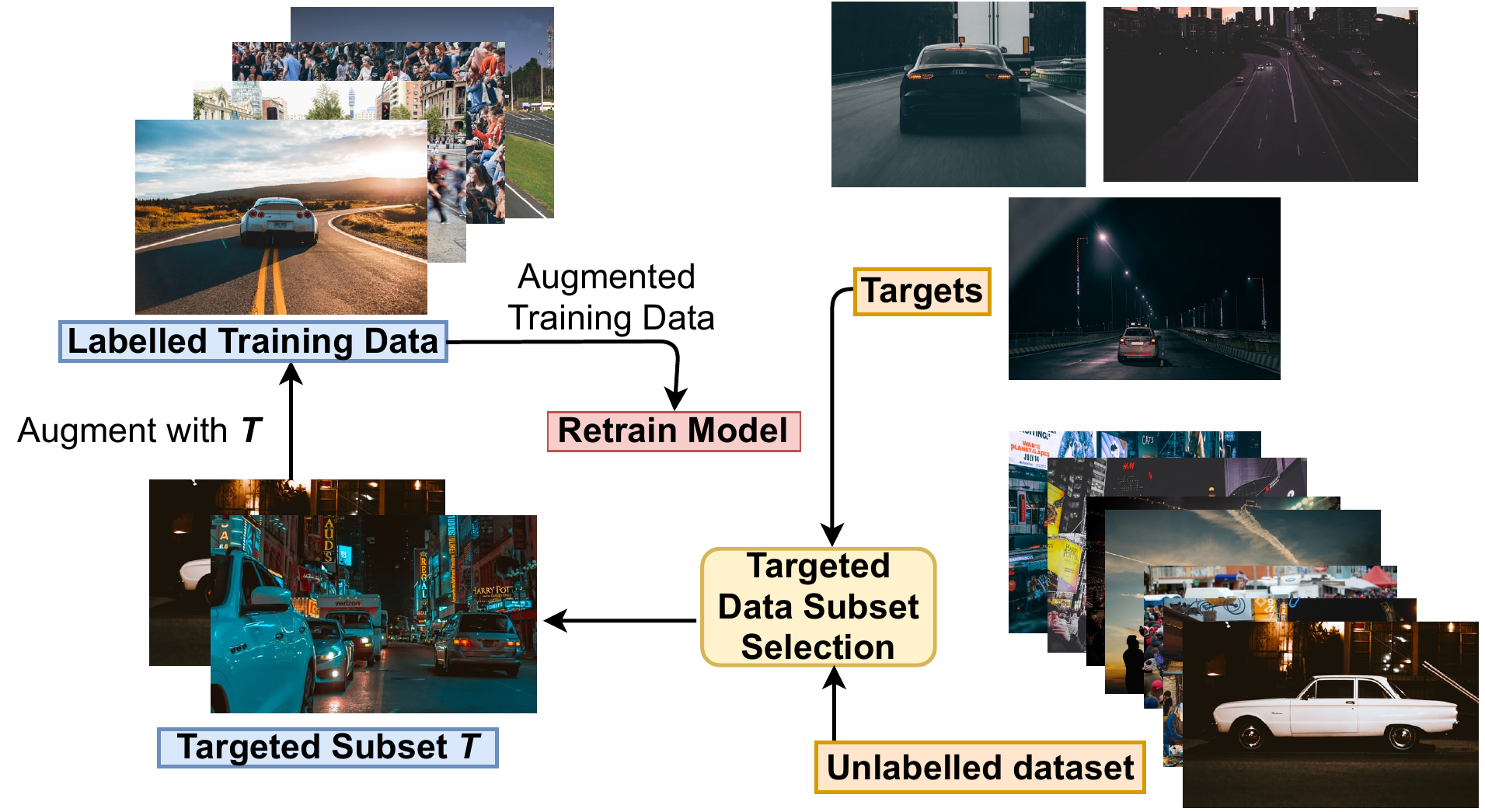}
\caption{Motivating example for targeted data subset selection (\modelss): the night images (target) are under-represented in training data. \modelss \vspace{0.5pt} mines for night images and augments the training data to improve the performance of the final model.}
\label{fig:tdss}
\end{wrapfigure}

\subsection{Targeted Data Subset Selection}
In real-world settings, the training data is often biased. Examples of such biases include distribution shift, imbalance in classes, presence of rare classes or rare slices, and out of distribution examples in the unlabeled dataset. In such cases a model's performance can be improved (at a given additional labeling cost) by augmenting the training data with some most informative samples matching the target distribution (hence called \emph{targeted} subset) from a large pool of unlabeled data. One way of achieving this is by assuming access to a clean validation set matching the target set distribution and using it as a target. Another example is where the target set is a critical slice of the data ({\em e.g.}, indoor images of people in the dark or images from specific classes that the user might care about) and we want to improve the model's performance on the target without sacrificing the overall accuracy and with minimum additional labeling costs (Fig. \ref{fig:tdss}). Yet another case of this is where the user is aware of a certain rare slice or class in the dataset, and has a few example images of this (say, either from the labeled set or from a held-out test set) of this rare slice. In this paper, we will study the problem of targeted subset selection to sample unlabeled data points. 

\section{Preliminaries}
\textbf{Submodular Functions: } We let $\Vcal$ denote the \emph{ground-set} of $n$ data points $\Vcal = \{1, 2, 3,...,n \}$ and a set function $f:
 2^{\Vcal} \xrightarrow{} \Re$.  
 The function $f$ is submodular~\citep{fujishige2005submodular}  if it satisfies the diminishing marginal returns, namely $f(j | \Xcal) \geq f(j | \Ycal)$ for all $\Xcal \subseteq \Ycal \subseteq \Vcal, j \notin \Ycal$.
Facility location, set cover, log determinants, {\em etc.} are some examples~\citep{iyer2015submodular}. Due to  close connections between submodularity and entropy, submodular functions can also be viewed as \emph{information functions}~\citep{zhang1998characterization}. Submodularity ensures that a greedy algorithm achieves bounded approximation factor when maximized~\citep{nemhauser1978analysis}.

\textbf{Submodular Mutual Information (MI):} Given a set of items $\Acal, \Bcal \subseteq \Vcal$, the submodular mutual information (MI)~\citep{levin2020online,iyer2020submodular} is defined as $I_f(\Acal; \Bcal) = f(\Acal) + f(\Bcal) - f(\Acal \cup \Bcal)$. Intuitively, this measures the similarity between $\Bcal$ and $\Acal$ and we refer to $\Bcal$ as the query set.

\cite{kaushal2020unified} extend MI to handle the case when the \emph{target} can come from an auxiliary set $\Vcal^{\prime}$ different from the ground set $\Vcal$. For targeted data subset selection, $\Vcal$ is the source set of data instances and the target is a subset of data points (validation set or the specific set of examples of interest).
Let $\Omega  = \Vcal \cup \Vcal^{\prime}$. We define a set function $f: 2^{\Omega} \rightarrow \Re$. Although $f$ is defined on $\Omega$, the discrete optimization problem will only be defined on subsets $\Acal \subseteq \Vcal$. To find an optimal subset given a query set $\Qcal \subseteq \Vcal^{\prime}$, we can define $g_{\Qcal}(\Acal) = I_f(\Acal; \Qcal)$, $\Acal \subseteq \Vcal$ and maximize the same.
\subsection{Examples of SMI functions}
We use the MI functions recently introduced in~\cite{iyer2020submodular, levin2020online} and their extensions introduced in \cite{kaushal2020unified}. 
For any two data points $i \in \Vcal$ and $j \in \Qcal$, let $s_{ij}$ denote the similarity between them.

\noindent\textbf{Graph Cut MI:} The submodular mutual information (SMI) instantiation of graph-cut (GCMI) is defined as: $I_{GC}(\Acal;\Qcal)=2\sum_{i \in \Acal} \sum_{j \in \Qcal} s_{ij}$.
Since maximizing GCMI maximizes the joint pairwise sum with the query set, it will lead to a summary similar to the query set $Q$. In fact, specific instantiations of GCMI have been intuitively used for query-focused summarization for videos ~\cite{vasudevan2017query} and documents ~\cite{lin2012submodularity, li2012multi}. 


\noindent\textbf{Facility Location MI - V1:} In the first variant of FL, we set $D$ to be $V$. The SMI instantiation of FL1MI can be defined as: $I_{FL1}(\Acal;\Qcal)=\sum_{i \in \Vcal}\min(\max_{j \in \Acal}s_{ij}, \eta \max_{j \in \Qcal}s_{ij})$.
The first term in the min(.) of FL1MI models diversity, and the second term models query relevance. An increase in the value of $\eta$ causes the resulting summary to become more relevant to the query.

\noindent\textbf{Facility Location MI - V2:} In the V2 variant, we set $D$ to be $V \cup Q$. The SMI instantiation of FL2MI can be defined as: $I_{FL2}(\Acal;\Qcal)=\sum_{i \in \Qcal} \max_{j \in \Acal} s_{ij} + \eta\sum_{i \in \Acal} \max_{j \in \Qcal} s_{ij}$.
FL2MI is very intuitive for query relevance as well. It measures the representation of data points that are the most relevant to the query set and vice versa. It can also be thought of as a bidirectional representation score.

\noindent\textbf{Log Determinant MI:} The SMI instantiation of LogDetMI can be defined as: $I_{LogDet}(\Acal;\Qcal)=\log\det(S_{\Acal}) -\log\det(S_{\Acal} - \eta^2 S_{\Acal,\Qcal}S_{\Qcal}^{-1}S_{\Acal,\Qcal}^T)$.
$S_{\Acal, \Qcal}$ denotes the cross-similarity matrix between the items in sets $\Acal$ and $\Qcal$. The similarity matrix in constructed in such a way that the cross-similarity between $\Acal$ and $\Qcal$ is multiplied by $\eta$ to control the trade-off between query-relevance and diversity.

\section{A Framework for Targeted Data Subset Selection}

We apply SMI functions to the setting of targeted data subset selection for improving a model's accuracy on some target classes/instances at a given additional labeling cost ($k$ instances) without compromising on the overall accuracy. Let $\Ecal$ be an initial training set of labeled instances and $\Tcal$ be the set of examples that the user cares about and desires better performance on. Let $\Ucal$ be a large unlabeled dataset. Using appropriate feature representation of the instances, we compute kernels of similarities of elements within $\Ucal$, within $\Tcal$ and between $\Ucal$ and $\Tcal$ to instantiate a MI function $I_f(\Acal; \Tcal)$ and maximize it to compute an optimal subset $\hat{\Acal} \subseteq \Ucal$ of size $k$ given $\Tcal$ as target (query) set. We then augment $\Ecal$ with labeled $\hat{\Acal}$ (i.e. $L(\hat{\Acal})$) and re-train the model to achieve the desired improvement. Through instantiating a rich class of MI functions including GCMI, FL1MI, FL2MI, COM and LogDetMI, \modelss\ offers a rich treatment to targeted subset selection. Our framework allows for adding an explicit diversity term $\gamma g(\Acal)$ where $\gamma$ is the weight and $g$ is a set function modeling diversity (for eg. total pairwise distance). This is helpful in cases when $I_f$ itself does not model diversity (for eg. GCMI). The algorithm is summarized in Algorithm~\ref{algo:tss}. Following ~\cite{ash2020deep, killamsetty2020glister} we use gradients as feature representation to compute the similarity kernels. The gradients are computed using model's inference for $\Ucal$ and $\Tcal$ and similarity is computed using cosine similarity. 

\begin{algorithm}
\begin{algorithmic}[1]
\REQUIRE Initial Labeled set of Examples: $\Ecal$, large unlabeled dataset: $\Ucal$, A target subset/slice where we want to improve accuracy: $\Tcal$, Loss function $\mathcal L$ for learning
\STATE Train model with loss $\mathcal L$ on labeled set $\Ecal$ and obtain parameters $\theta_E$
\STATE Compute the gradients $\{\nabla_{\theta_E} \mathcal L(x_i, y_i), i \in \Ucal\}$ and $\{\nabla_{\theta_E} \mathcal L(x_i, y_i), i \in \Tcal\}$.
\STATE Using the gradients, compute the similarity kernels  and define a submodular function $f$ and diversity function $g$
\STATE $\hat{\Acal} \gets \max_{\Acal \subseteq \Ucal, |\Acal|\leq k} I_f(\Acal;T) + \gamma g(\Acal)$
\STATE Obtain the labels of the elements in $\Acal^*$: $L(\hat{\Acal})$
\STATE Train a model on the combined labeled set $\Ecal \cup L(\hat{\Acal})$
\end{algorithmic}
\caption{\modelss}
\label{algo:tss}
\end{algorithm}



\section{Effectiveness of SMI for \modelss} \label{subsec:exp-tss}

\textbf{Dataset, Baselines and Implementation details: } 
We demonstrate the effectiveness of \modelss\ in obtaining a targeted subset for improving image classification accuracy for some target classes on CIFAR-10 and MNIST datasets. To simulate a real-world setting, 
we split the available train set into train, validate and a data lake such that (i) the train set has few labeled instances and poorly represents two randomly picked classes (target), and (ii) data lake is a large set whose labels we do not use (resembling a large pool of unlabeled data in real-world). The poorly represented classes do not perform well on the validation set and hold clue to picking up the target of interest. Performance is measured on the test set from the respective datasets. We then apply \modelss\ (Algorithm~\ref{algo:tss}) comparing MI functions with other existing approaches. Specifically, for MI functions we use LogDetMI, GCMI, FL1MI, FL2MI, and GCMI + Diversity (equivalent to an intuitive approach of minimizing average gradient difference with the target)
For existing approaches, we compare with three active learning baselines (uncertainty sampling (US), \textsc{Badge}, and \textsc{Glister-Active} (GLISTER)) running them only once as per our setting (i.e. we select the unlabeled subset only once). Since these active learning baselines do not explicitly have information of the target set, to further strengthen them we also compare against two variants which are target-aware. The first is `targeted uncertainty sampling' (TUS) where a product of the uncertainty and the similarity with the target is used to identify the subset, and second is \textsc{Glister-TSS}
where the target set is used in the bi-level optimization. Finally, we also compare with pure diversity/representation functions (Facility Location (FL), Graph Cut (GC), Log Determinant (LogDet), Disparity-Sum (DSUM)) and random sampling. We train the model (ResNet-18~\citep{he2016deep} for CIFAR-10, LeNet~\citep{lecun1989backpropagation} for MNIST) using cross-entropy loss and SGD optimizer until training accuracy exceeds 99\% (Base model). After augmenting the train set with the labeled version of the selected subset and re-training the model, we report the average gain in accuracy for the target classes and overall gain in accuracy across all classes on test set, averaged across 10 runs of randomly picking any two classes as target. We run \modelss\ for different budgets and also study the effect of budget on the performance. Wherever applicable, we keep the internal parameters at their default values of 1. 

\begin{figure}
    \centering 
  \includegraphics[width=1.0\textwidth]{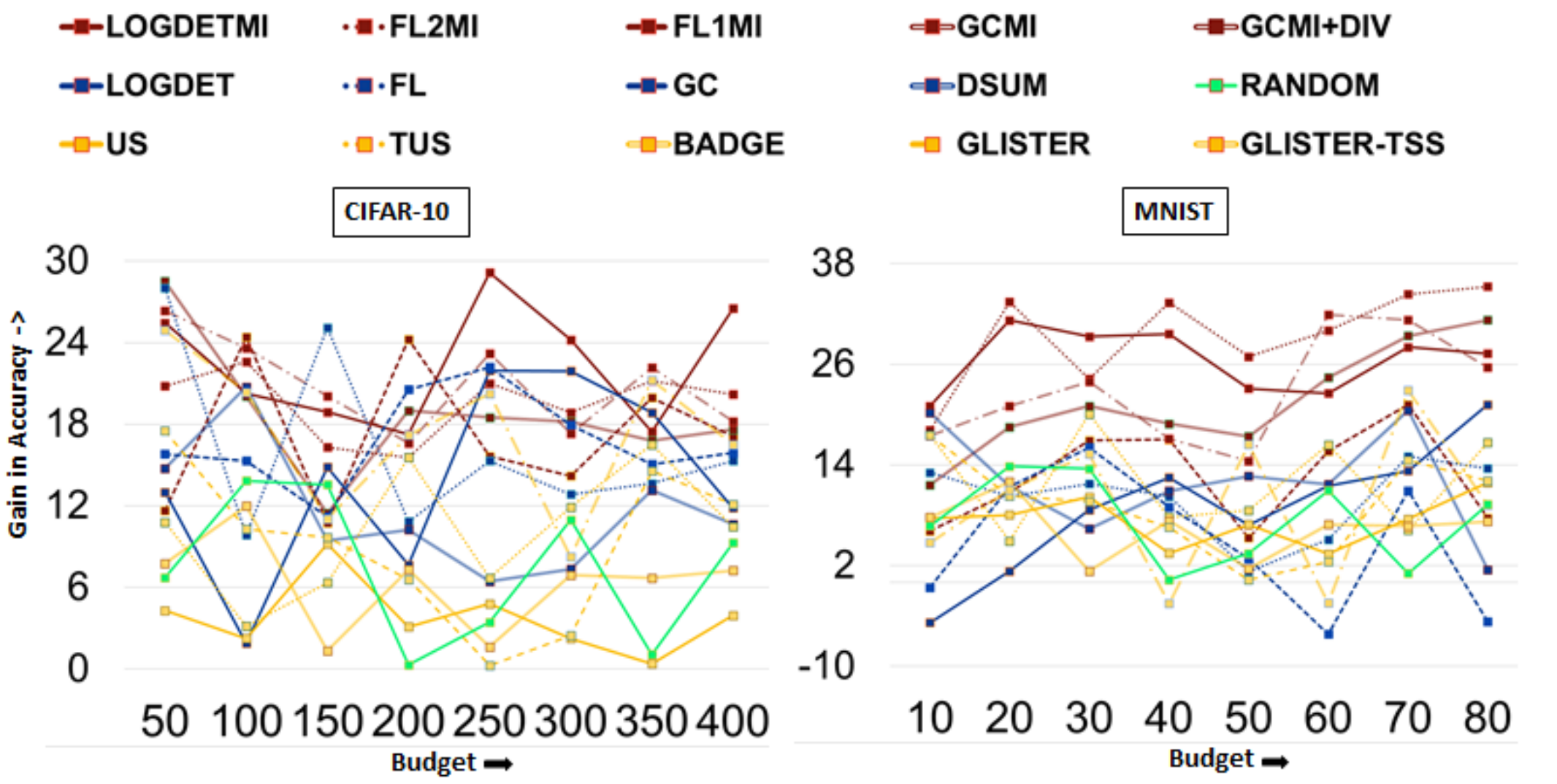}
\caption{Comparison of different methods for targeted subset selection for different budgets on CIFAR-10 and MNIST. X-axis: budgets, Y-axis: gain in model accuracy for target classes on test set. MI based approaches (lines in {\textcolor{red}{red})} significantly outperform others across all subset sizes. (Section~\ref{subsec:exp-tss}).}

\label{fig:gain-size}
\end{figure}

\textbf{Results: } In Table~\ref{tab:cifar-mnist-results}, we report the results for a budget of 400 for CIFAR-10 and 70 for MNIST. To keep the setting as realistic as possible, we set the target set to be much smaller than the budget (around 10\% of the budget -- 10 for CIFAR-10 and 6 for MNIST). We report the effect of budget on the gain in accuracy of the target classes in Fig.~\ref{fig:gain-size}. On both datasets, MI functions yield the best improvement in accuracy on the target classes ($\approx$ 20-30\% gain over the model's performance before re-training with added targeted subset; $\approx$ 12\% more than other methods) while also simultaneously increasing the overall accuracy by $\approx$ 2-6\%. They consistently outperform \textsc{Badge}, \textsc{Glister-TSS}, \textsc{US} and \textsc{TUS} across all budgets. 
Since the SMI functions (LogDetMI, Fl2MI and GCMI+DIV) model both query-relevance and diversity, they perform better than both a) functions which tend to prefer relevance (GCMI, TUS) and b) functions which tend to prefer diversity/representation (\textsc{Badge}, FL, GC, DSUM, LogDet). 
Also, we observe that across different budgets, the MI functions outperform other methods by greater margins on the target class accuracy (Fig.~\ref{fig:gain-size}).
This is expected, as other methods are not effective in considering the target. 

\begin{table}[ht]
\begin{center}
\begin{tabular}{lrrrr}
\hline
Method               & \multicolumn{2}{c}{CIFAR-10}                                     & \multicolumn{2}{c}{MNIST}                                        \\ \hline
                     & \multicolumn{1}{l}{Target} & \multicolumn{1}{l}{Overall} & \multicolumn{1}{l}{Target} & \multicolumn{1}{l}{Overall} \\ \cline{2-5}
Base            & 11.2                               & 42.2                        & 52.76                               & 86.8 
\\
\hline
Random      & +2.75  & +1.43 & +1.08   & -0.032 \\
BADGE \citep{ash2020deep}       & +7.245 & +2.38 & +6.7    & +1.659 \\
GLISTER \citep{killamsetty2020glister}     & +12.1  & +2.27 & +14.56  & +2.27  \\
GLISTER-TSS & +16.5  & +1.78 & +22.895 & +4.05  \\
US \citep{settles2009active}          & +3.95  & +2.03 & +7.56   & +1.182 \\
TUS         & +10.45 & \textcolor{red}{+2.99} & +6.21   & +1.611 \\
LogDet      & +11.85 & +1.2  & +13.29  & +1.89  \\
FL          & +15.3  & \textcolor{green}{+2.63} & +15.025 & +2.41  \\
GC          & +15.9  & +1.79 & +10.935 & +1.16  \\
DSUM        & +10.65 & +1.9  & +20.515 & +3.92  \\ \hline
LogDetMI    & \textcolor{blue}{+26.5}  & +2.21 & +28.035 & \textcolor{blue}{+5.26}  \\
FL2MI       & \textcolor{red}{+20.2}  & +1.7  & \textcolor{blue}{+34.36}  & \textcolor{green}{+5.14}  \\
FL1MI       & +17.1  & +2.28 & +21.21  & +3.83  \\
GCMI        & +17.6  & +1.48 & \textcolor{green}{+29.375} & \textcolor{red}{+5.21}  \\
GCMI+DIV   & \textcolor{green}{+18.2}  & \textcolor{blue}{+3.74} & \textcolor{red}{+31.28}  & +4.21 \\

\hline
\end{tabular}

\caption{Comparison of \modelss\ (MI functions) with other methods for a budget of 400 (CIFAR-10) and 70 (MNIST). The numbers are the gain in \% accuracy of the target classes (Target) and all classes (Overall) over the Base model after re-training the model (see text). Highest in \textcolor{blue}{blue}, $2^{nd}$ and $3^{rd}$ highest in \textcolor{red}{red} and \textcolor{green}{green} respectively. }
\label{tab:cifar-mnist-results}
\end{center}
\end{table}

\section{Conclusion}
We demonstrate the effectiveness of SMI functions for improving a model's performance by augmenting the training data with samples that match a target distribution (targeted data subset selection). Through experiments on CIFAR-10 and MNIST datasets, we empirically verify the superiority of SMI functions over existing methods.
\bibliography{main}


\end{document}